%% file: main.tex
\def\year{2022}\relax
\documentclass[letterpaper]{article} 
\usepackage{aaai22}  
\usepackage{times}  
\usepackage{helvet}  
\usepackage{courier}  
\usepackage[hyphens]{url}  
\usepackage{graphicx} 
\usepackage{natbib}  
\usepackage{caption} 
\usepackage{mathtools}

\DeclareCaptionStyle{ruled}{labelfont=normalfont,labelsep=colon,strut=off} 
\usepackage{algorithm}
\usepackage{algorithmic}
\usepackage{color}

\usepackage{float}
\usepackage{newfloat}
\usepackage{listings}
\floatstyle{ruled}
\newfloat{listing}{tb}{lst}{}
\floatname{listing}{Listing}

\usepackage{graphicx}
\usepackage{subfigure}
\usepackage{multirow}
\usepackage{tcolorbox}
\usepackage{amsmath}

\DeclareMathOperator*{\concat}{%
    \mathchoice%
        {\Big\Vert}%
        {\big\Vert}%
        {\Vert}%
        {\Vert}%
}

\title{Towards Relation-centered Pooling and Convolution for Heterogeneous Graph Learning Networks}
\author {
    Tiehua Zhang\textsuperscript{\rm 1}\thanks{ co-first author (corresponding author: Tiehua Zhang).}, 
    Yuze Liu\textsuperscript{\rm 1}$^{*}$, 
    Yao Yao\textsuperscript{\rm 1}, 
    Youhua Xia\textsuperscript{\rm 2}, 
    Xin Chen\textsuperscript{\rm 1}, 
    Xiaowei Huang\textsuperscript{\rm 1}, 
    Jiong Jin\textsuperscript{\rm 3}
}
\affiliations {
    \textsuperscript{\rm 1} Ant Group\\
    \textsuperscript{\rm 2} School of Computer Science, Wuhan University\\
    \textsuperscript{\rm 3} School of Science, Computing and Engineering Technologies, Swinburne University of Technology\\
    \{zhangtiehua.zth,liuyuze.liuyuze,yy350608,jinming.cx,wei.huangxw\}@antgroup.com, 
    xiayouhua@whu.edu.cn
    jiongjin@swin.edu.au
}

\usepackage{bibentry}

\begin{document}

\maketitle

\input{0_abstract.tex}

\input{1_introduction.tex}

\input{2_related_work.tex}

\input{3_methdology.tex}

\input{4_experiment.tex}

\input{5_conclusion.tex}

\bibliography{aaai22.bib}

\end{document}

%% file: 0_abstract.tex
\begin{abstract}
Heterogeneous graph neural network has unleashed great potential on graph representation learning and shown superior performance on downstream tasks such as node classification and clustering. Existing heterogeneous graph learning networks are primarily designed to either rely on pre-defined meta-paths or use attention mechanisms for type-specific attentive message propagation on different nodes/edges, incurring many customization efforts and computational costs. To this end, we design a relation-centered \textbf{P}ooling and \textbf{C}onvolution for \textbf{H}eterogeneous \textbf{G}raph learning \textbf{N}etwork, namely PC-HGN, to enable relation-specific sampling and cross-relation convolutions, from which the structural heterogeneity of the graph can be better encoded into the embedding space through the adaptive training process. We evaluate the performance of the proposed model by comparing with state-of-the-art graph learning models on three different real-world datasets, and the results show that PC-HGN consistently outperforms all the baseline and improves the performance maximumly up by 17.8\%.
\end{abstract}

%% file: 1_introduction.tex
\section{Introduction}
There has been increasing research on modelling graph data as such data becomes more ubiquitous in real-world scenarios. These well-organized graph data contain rich yet complex semantic and structural information of various real-world systems such as recommendation~\cite{shi2018heterogeneous}, medical diagnosis~\cite{rong2020self, fedrel} and text analysis~\cite{bahdanau2014neural}. To better leverage such information, graph representation learning, which aims to learn a function that maps the input space into the low-dimensional embedding space while preserving the attributes of the graph, aroused to the spotlight to serve the purpose of node/graph classification and clustering tasks. Conventionally, matrix factorization (e.g., adjacency matrix) methods~\cite{newman2006modularity,weisfeiler1968reduction} have been used to learn the latent embedding vectors of a graph, yet it suffers from both the computational cost on large-scale graph data and statistical performance drawback~\cite{shi2016survey}. 

Graph Neural Networks (GNN) have recently become the most prevalent technique to enhance graph learning performance. However, most GNN models, such as GCN, GraphSAGE, GAT and GPS, are initially designed for homogeneous graphs~\cite{wang2022survey}, which only involve one type of node and edge. These models thus cannot be directly applicable to graphs with different types of entities (i.e., heterogeneous graphs with various node/edge types). Specifically, heterogeneous graphs present a more complex semantic-dependent topology, meaning that the local structure of one node varies in different types of relations. Apart from that, in heterogeneous graphs, there are a number of types of nodes and edges, each of which represents a particular type of attribute in the graph and thus requires different feature spaces. The heterogeneity attributes need to be taken into account when designing heterogeneous embedding methods, as it is essential to ponder on the significant roles of nodes and edges altogether. Currently, much research focuses on designing the specialized GNNs to learn representations in heterogeneous graphs. For instance, both HetGNN~\cite{zhang2019heterogeneous} and HAN~\cite{han} rely on pre-defined meta-paths to perform graph learning. Other existing works either emphasize on capturing the characteristics of nodes (e.g., HetSANN~\cite{hetsann}, HGT~\cite{hgt}) or multi-typed edge properties (e.g., RGCN~\cite{RGCN}, MBGCN~\cite{MBGCN}). While the most recent work R-HGNN~\cite{rhgnn} starts to learn both the semantic meaning of edge and node features in a collaborative manner, it incurs a high computational cost due to the heavy use of attentions at both node and edge level.

In light of these limitations, we propose an adaptive relation-centric heterogeneous graph learning method, aiming to better incorporate the information of multi-relation types without concerning pre-defined meta-paths in heterogeneous graphs. Specifically, we first design the relation-specific pooling to enable the importance-based sampling, aiming to only select a subset of neighbor nodes to improve the learning efficiency. Based on the sampled neighbors of each node, a cross-relation convolution is implemented to learn the heterogeneity of the graph through different types of relations. After that, both the pooling and convolution can be optimized end-to-end. The contributions are summarized as follows:

\begin{itemize}
\item We propose a novel graph learning model to enable both relation-specific pooling and cross-relation convolutions on heterogeneous graphs. The model can be directly applied to any heterogeneous graphs without concerning the pre-defined meta-paths.
\item We project relation-specific neighbors into the same latent space regardless of the neighbor types and design an adaptive importance-based pooling to enable sampling for each relation type. Benefiting from the sampled neighbors set, an effective cross-relation convolution is implemented to incorporate the heterogeneity attributes of the graph into the node embeddings.
\item We conduct extensive experiments to evaluate the performance of the proposed model. The results show that PC-HGN excels in all three real-world datasets compared with state-of-the-art homogeneous and heterogeneous graph learning models. We also analyze the impact of filters and hidden node dimensions to interpret the effectiveness of the proposed model.
\end{itemize}

%% file: 2_related_work.tex
\section{Related Work}

There has been a rich line of research on graph representation learning in recent years. The graph convolution network (GCN)~\cite{gcn} is the first of its kind on learning the node embedding through graph Laplacian methods. Following that, graph attention network (GAT)~\cite{gat} introduces multi-head attention mechanisms on each edge to enable the attentive message aggregation when updating the node embeddings. As opposed to aggregating the message from all connected nodes like GCN and GAT, researchers start to focus on designing more efficient graph learning algorithms. Several sampling-based graph learning methods, including GraphSAGE~\cite{graphsage}, AS-GCN~\cite{asgcn}, FastGCN~\cite{fastgcn}, GPS~\cite{gps}, were developed for fast representation learning on graphs. To be more specific, both GraphSAGE, and GPS adopt the node-based sampling, and it considers only part of the connected neighbor nodes when aggregating the neighbor messages. AS-GCN and FastGCN, on the other hand, design different layer-based sampling strategies. While FastGCN constructs each layer from sampled nodes independently, AS-GCN samples all neighborhoods nodes altogether and allow the neighborhood sharing, in which the lower layer is sampled conditionally on the top one to preserve the between-layer connections. GraphSAINT~\cite{graphsaint} proposes a graph-level sampling approach, in which a full graph is downsampled into subgraphs based on designated node, edge and random walk samplers. Similarly, some research explore the pooling operations on graph. For instance,~\cite{defferrard2016convolutional} uses a binary tree indexing for graph coarsening, in which 1-D pooling operations are applied on the fixed node indices in the tree. ~\cite{ying2018hierarchical} achieves the pooling with the help of an assignment matrix, and nodes can be assigned to different clusters of the next layer. Graph U-Nets designs a encoder-decoder pooling architecture to speed up the large-scale graph learning. The gPool encoding layer adaptively selects nodes according to the scalar projection value on the trainable projection vector to form a small subgraph, and the inverse gUnpool decoding layer is used to restore the original structure of the graph by using the position information of the selected nodes. 

As pointed out in~\cite{wang2020survey}, the sampling on heterogeneous graph is still in its infancy stage and worth exploring owing to the rich structure and semantic information across different node/edge types. Two scalable representation learning models, namely metapath2vec and metapath2vec++, are developed~\cite{dong2017metapath2vec}. Specifically, metapath2vec builds heterogeneous neighborhoods of nodes based on random walks on the pre-defined metapaths, and uses heterogeneous skip-gram model to perform node embedding. Metapath2vec++ further explores synchronous modeling of structural and semantic correlations across different edges in heterogeneous graphs. Recently, many efforts aim to develop GNN-related models for heterogeneous graph learning. HAN~\cite{han} aims to learn the importance between a node and its meta-path based neighbors, and semantic-level attention is designed to learning the hand-designed mete-paths. Following that, type-specific parameters are considered in HGT~\cite{hgt} to learn the characteristics of different nodes and relations. To make the training on web-scale data more efficient, it designs a HGSampling algorithm to sample an equal number of nodes per node type during training process. R-HGNN~\cite{rhgnn} proposes a relation-aware method to learn the semantic representation of relations while discerning the node representations with respect to different relation types.

%% file: 3_methdology.tex
\begin{figure*}[h!]
    \vspace{-6mm}
    \centering
    \includegraphics[width =\textwidth]{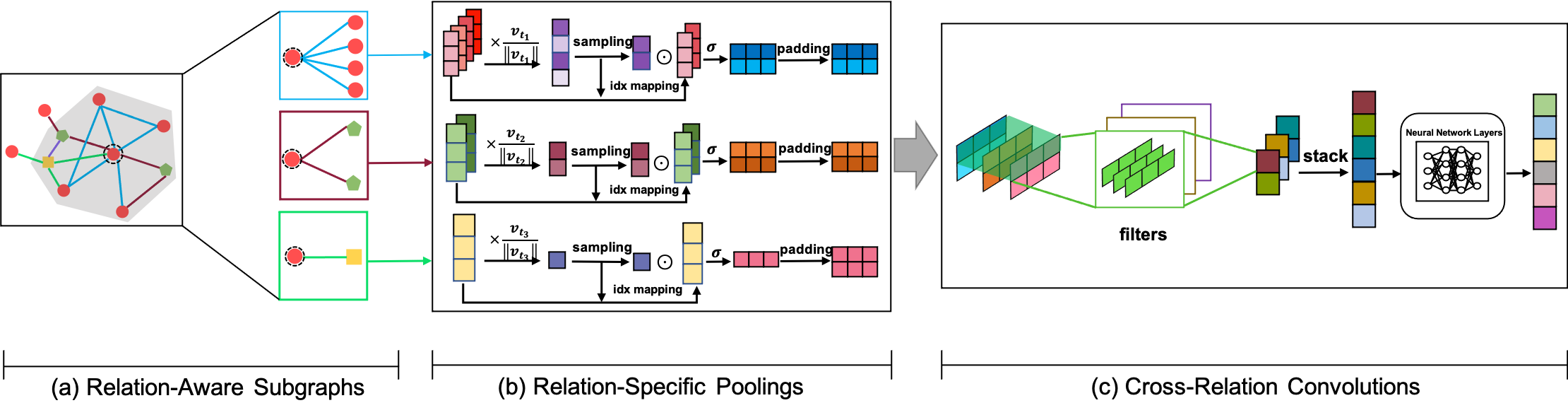}
    \caption{The overall architecture of PC-HGN.}
    \vspace{-6mm}
    \label{fig:demo}
\end{figure*}
\section{Methodology}
We present the proposed PC-HGN in this section. We first describe the notations that denote the heterogeneous graphs and then elaborate on relation-specific pooling and cross-relation convolutions, respectively.
\subsection{Heterogeneous Graph Definition}
We formally define the heterogeneous graph in this part. Given a heterogeneous graph $\mathcal{G} = \left(\mathcal{V},\mathcal{E},\mathcal{T}_{v},\mathcal{T}_{e}\right)$, it encompasses a number of nodes $\mathcal{V}$ with multiple node types $\mathcal{T}_{v}$ and edges $\mathcal{E}$ with multiple edge types $\mathcal{T}_{e}$. We define the node type mapping function $\phi:\mathcal{V}\rightarrow\mathcal{T}_{v}$ and the edge type mapping function $\psi:\mathcal{E}\rightarrow\mathcal{T}_{e}$, respectively~\cite{sun2013mining}. 
Each heterogeneous graph presents multiple edge types, meaning $\left|\mathcal{T}_{e}\right|>1$. We use $N$ to denote the number of nodes in the graph $\left|\mathcal{V}\right| = N$, and the neighbor of each node $v$ is represented as $\mathcal{N}\left(v\right)$. Simply, $\emph{e}_{uv}$ denotes the edge connection from $u$ to $v$. We use $\emph{X}\in\mathcal{R}^{N\times D}$  to represent the initial node feature matrix, and node $v$ 's initial feature vector is $\emph{x}_{v}\in\mathcal{R}^{D}$.

\subsection{Relation-specific Pooling}
As pointed out in~\cite{lee2019self}, pooling operation is considered effective to improve both training efficiency and generalization ability of the model. To enable the pooling on semantic-dependent heterogeneous graphs, we propose the relation-specific pooling operations, which enables an adaptive sampling on neighbor nodes for each relation type.

As illustrated in Fig.~\ref{fig:demo} (a), we first split the graph into different subgraphs based on different relations.
Given a target node $v$ and its neighbor nodes $\mathcal{N}\left(v\right)$, we define a relation as $\left<\phi\left(u\right),\psi\left(\emph{e}_{uv}\right),\phi\left(v\right)\right>$ between node $\emph{v}$ and node $u\in\mathcal{N}\left(v\right)$, indicating
there exists an edge $\emph{e}_{uv}$ from neighbor node $u$ to target node $v$ with edge type $\psi\left(\emph{e}_{uv}\right)$. Then, We conduct relation-specific pooling operation on each subgraph to get sampled neighbor nodes under that relation. Specifically, for any node $v$, its neighbor node set can be represented as $\mathcal{N}_{v}^{\emph{t}} = \left\{u\big|u\in\mathcal{N}\left(v\right),\psi\left(\emph{e}_{uv}\right) = t,t\in\mathcal{T}_{e},v\in\mathcal{V}\right\}$, in which all nodes belongs to only one relation. The initial node feature matrix is then denoted as $\emph{X}_{\mathcal{N}_{v}}^t$.
We formulate the relation-specific pooling of $t\in\mathcal{T}_{e}$ in the following:
\begin{equation}
  \begin{aligned}
    \emph{s}^{t}_{\mathcal{N}_{v}} &= \emph{X}_{\mathcal{N}_{v}}^t\cdot\frac{\emph{v}_{t}}{\left\|\emph{v}_{t}\right\|}, \\
    \emph{idx}_{\mathcal{N}_{v}}^{t} &= \emph{rank}\left(\emph{s}^{t}_{\mathcal{N}_{v}},k\right),\\
    \hat{\emph{s}}^{t}_{\mathcal{N}_{v}} &= \sigma\left(\emph{s}^{t}_{\mathcal{N}_{v}}\left[\emph{idx}_{\mathcal{N}_{v}}^{t}\right]\right),\\
    \hat{\emph{X}}^{t}_{\mathcal{N}_{v}} &= \emph{X}_{\mathcal{N}_{v}}^t \left[\emph{idx}_{\mathcal{N}_{v}}^{t},:\right], \\
    \Tilde{\emph{X}}_{\mathcal{N}_{v}}^t &= \hat{\emph{X}}^{t}_{\mathcal{N}_{v}}\odot\left(\hat{\emph{s}}^{t}_{\mathcal{N}_{v}}\cdot\textbf{1}_{1\times D}\right)   
  \end{aligned}
\end{equation}

For each relation, we project the neighbors under that relation to a latent space via a trainable  vector $\emph{v}_{t}$, from which the respective important scores are calculated as $\emph{s}^{t}_{\mathcal{N}_{v}}\in\mathcal{R}^{\left|\mathcal{N}^t_v\right|}$. In other word, it measures how much information of neighbor nodes can be retained when projected onto the direction of $\emph{v}_t$ for each relation $t$.
$\emph{rank}\left(\emph{s}^{t}_{\mathcal{N}_{v}},k\right)$ is the importance-based node ranking function,  which returns indices of the $k$-largest values in $\emph{s}^{t}_{\mathcal{N}_{v}}$.
$\emph{idx}_{\mathcal{N}_{v}}^{t}$, returned by the ranking function, contains the indices of neighbors selected for target node $v$.
$\emph{s}^{t}_{\mathcal{N}_{v}}\left[\emph{idx}_{\mathcal{N}_{v}}^{t}\right]$ extracts values with indices, followed by a non-linear function $\sigma$ to get weighted score vector $ \hat{\emph{s}}^{t}_{\mathcal{N}_{v}}$.
Similarly, $\emph{X}_{\mathcal{N}_{v}}^t\left[\emph{idx}_{\mathcal{N}_{v}}^{t},:\right]$ perform the row extraction to form the feature matrix of selected neighbors, whose result is denoted as $\hat{\emph{X}}^{t}_{\mathcal{N}_{v}}$.
$\textbf{1}_{1\times D}\in\mathcal{R}^{1\times D}$ is a row vector with all elements being $1$, and $\odot$ denotes the element-wise matrix multiplication. It is worth noting that in case of lack of neighbor nodes in a specific relation, we apply mean operation on $\hat{\emph{X}}^{t}_{\mathcal{N}_{v}}$ for padding purpose, guaranteeing $\Tilde{\emph{X}}_{\mathcal{N}_{v}}^{t}\in\mathcal{R}^{k\times D}$.

\subsection{Cross-relation Convolutions}
The process of cross-relation convolution aims to consider both the heterogeneity attributes and node feature embedding simultaneously, which is shown in Fig~\ref{fig:demo} (c).
The output of relation-specific pooling is represented as $\mathcal{X}_{\mathcal{N}_{v}} = \left[\Tilde{\emph{X}}_{\mathcal{N}_{v}}^{t_{1}},\Tilde{\emph{X}}_{\mathcal{N}_{v}}^{t_{2}},...,\Tilde{\emph{X}}_{\mathcal{N}_{v}}^{t_{\left|\mathcal{T}_{e}\right|}}\right]$, which contains sampled neighbor node features $\Tilde{\emph{X}}_{\mathcal{N}_{v}}^{t}\in\mathcal{R}^{k\times D}$ for each relation $t$. Therefore, the sampled feature matrices with different relations are stacked into the three-dimensional tensor $\mathcal{X}_{\mathcal{N}_{v}}\in\mathcal{R}^{\left|\mathcal{T}_{e}\right|\times k\times D}$. Note that for computation convenience, we apply zero-padding along the relation dimension if such relation does not exist for node $v$.
When it comes to convolution, we use $P$ trainable filters, each of which encompasses different kernels stacked together along the relation dimension $\mathcal{K}^{p} = \left[\emph{K}_p^{t_{1}},\emph{K}_p^{t_{2}},...,\emph{K}_p^{t_{\left|\mathcal{T}_{e}\right|}}\right] \in\mathcal{R}^{\left|\mathcal{T}_{e}\right|\times s\times D}, p \in \left[1,...,P\right]$.
Each kernel $\emph{K}_{p}^{t}\in\mathcal{R}^{s\times D}$ in $p$-th filter is used to generate the feature map of relation $t$, where $s$ denotes the number of search window.
The process of cross-relation convolution of filter $p$ is formulated as:
\begin{equation}
    \emph{h}_{v}^{p} = \mathcal{X}_{\mathcal{N}_{v}} \circ \mathcal{K}^{p}
\end{equation}
where $\emph{h}_{v}^{p}\in\mathcal{R}^{\left(k-s+1\right)}$ represents the convoluted results.
Concretely, the $m$-th element of $\emph{h}_{v}^{p}$ is calculated as:
\begin{equation}
    \left(\mathcal{X}_{\mathcal{N}_{v}} \circ \mathcal{K}^{p}\right)_{\left(m\right)} = \sum_{i = 1}^{\left|\mathcal{T}_{e}\right|}\sum_{j = 1}^{s}\Tilde{\emph{X}}^{t_{i}}_{\mathcal{N}_{v}}\left[m+j-1,:\right]\cdot\left(\emph{K}^{t_{i}}_{p}\left[j,:\right]\right)^{T}
\end{equation}
where $\Tilde{\emph{X}}^{t_{i}}_{\mathcal{N}_{v}}\left[m+j-1,:\right]$ and $\emph{K}^{t_{i}}_{p}\left[j,:\right]$ indicates the index dependent row extractions. Following that, the convoluted results of each filter are stacked to generate the input of a shallow multi-layer perceptions (MLP), represented as:
\begin{equation}
    \hat{\emph{h}}_{v} = \emph{MLP}\left(\concat_{p=1}^P\emph{h}^{p}_{v}\right)
\end{equation}

Herein, $\hat{\emph{h}}_{v}$ represents the learnt hidden embedding of node $v$.

%% file: 4_experiment.tex
\section{Experiment}
In this section, we conduct extensive experiments, aiming to answer the following research questions: 
\begin{itemize}
    \item \textbf{RQ1:} how does our method perform compared with other state-of-the-art homogeneous/heterogeneous baselines for representation learning.
    \item \textbf{RQ2:} what is the impact of cross-relation convolutions on the model.
    \item \textbf{RQ3:} how much the hyper-parameters (e.g., embedding dimension and the number of sampled heterogeneous neighbors) affect the performance of the model.
\end{itemize}
 
\begin{table*}[t!]
\label{tab:results}
\resizebox{\textwidth}{!}{
\begin{tabular}{|cc|llllll|}
\hline
\multicolumn{2}{|c|}{\multirow{3}{*}{Methods}}                  & \multicolumn{6}{c|}{Dataset}                                                                                                                           \\ \cline{3-8} 
\multicolumn{2}{|c|}{}                                          & \multicolumn{2}{c|}{DBLP}                               & \multicolumn{2}{c|}{ACM}                                & \multicolumn{2}{c|}{IMDB}          \\ \cline{3-8} 
\multicolumn{2}{|c|}{}                                          & \multicolumn{1}{l|}{F1-Micro} & \multicolumn{1}{l|}{F1-Macro} & \multicolumn{1}{l|}{F1-Micro} & \multicolumn{1}{l|}{F1-Macro} & \multicolumn{1}{l|}{F1-Micro} & F1-Macro \\ \hline
\multicolumn{1}{|c|}{\multirow{4}{*}{Homogeneous Models}}   & GCN       & \multicolumn{1}{l|}{0.925}      & \multicolumn{1}{l|}{0.916}      & \multicolumn{1}{l|}{0.879}      & \multicolumn{1}{l|}{0.880}      & \multicolumn{1}{l|}{0.518}      &  \multicolumn{1}{l|}{0.494}     \\ \cline{2-8} 
\multicolumn{1}{|c|}{}                              & GAT       & \multicolumn{1}{l|}{0.909}      & \multicolumn{1}{l|}{0.900}      & \multicolumn{1}{l|}{0.897}      & \multicolumn{1}{l|}{0.898}      & \multicolumn{1}{l|}{0.545} & \multicolumn{1}{l|}{0.513}           \\ \cline{2-8} 
\multicolumn{1}{|c|}{}                              & GraphSAGE & \multicolumn{1}{l|}{0.926}      & \multicolumn{1}{l|}{0.915}      & \multicolumn{1}{l|}{0.916}      & \multicolumn{1}{l|}{0.917}      & \multicolumn{1}{l|}{0.554}      &     \multicolumn{1}{l|}{0.539}   \\ \cline{2-8} 
\multicolumn{1}{|c|}{}                              & GraphSAINT       & \multicolumn{1}{l|}{0.923}      & \multicolumn{1}{l|}{0.912}      & \multicolumn{1}{l|}{0.908}      & \multicolumn{1}{l|}{0.909}      & \multicolumn{1}{l|}{0.535}      &   \multicolumn{1}{l|}{0.504}    \\ \hline\hline
\multicolumn{1}{|c|}{\multirow{4}{*}{Heterogeneous Models}} & HAN       & \multicolumn{1}{l|}{0.934}      & \multicolumn{1}{l|}{0.925}      & \multicolumn{1}{l|}{0.920}      & \multicolumn{1}{l|}{0.921}      & \multicolumn{1}{l|}{0.548}      &  \multicolumn{1}{l|}{0.530}      \\ \cline{2-8} 
\multicolumn{1}{|c|}{}                              & HGT       & \multicolumn{1}{l|}{0.838}      & \multicolumn{1}{l|}{0.824}      & \multicolumn{1}{l|}{0.898}      & \multicolumn{1}{l|}{0.897}      & \multicolumn{1}{l|}{0.556}      &    \multicolumn{1}{l|}{0.408}    \\ \cline{2-8} 
\multicolumn{1}{|c|}{}                              & R-HGNN     & \multicolumn{1}{l|}{0.930}      & \multicolumn{1}{l|}{0.919}      & \multicolumn{1}{l|}{0.916}      & \multicolumn{1}{l|}{0.917}      & \multicolumn{1}{l|}{0.561}      &   \multicolumn{1}{l|}{0.543}      \\ \cline{2-8} 
\multicolumn{1}{|c|}{}                              & PC-HGN      & \multicolumn{1}{l|}{\textbf{0.938}}      & \multicolumn{1}{l|}{\textbf{0.929}}      & \multicolumn{1}{l|}{\textbf{0.928}}      & \multicolumn{1}{l|}{\textbf{0.926}}      & \multicolumn{1}{l|}{\textbf{0.599}}      &  \multicolumn{1}{l|}{\textbf{0.581}}      \\ \hline
\end{tabular}}
\caption{Comparisons with baseline models on node classification task}
\label{tab:performance}
\end{table*}

\subsection{Dataset}
We adopt the the following three real-world datasets to verify the performance of the proposed model, including two citation networks DBLP and ACM, and a moive dataset ACM.
\begin{itemize}
    \item \textbf{DBLP}\footnote{https://s3.cn-north-1.amazonaws.com.cn/dgl-data/dataset/openhgnn/dblp4GTN.zip} contains three different types of nodes, including Paper(P), Author(A), and Conference(C). Also, edges are presented as P-A, A-P, P-C and C-P, each representing a respective edge type. Four different categories of papers are defined as labels of this dataset. The initial node features are calculated using bag-of-words. 
    \item \textbf{ACM}\footnote{ https://s3.cn-north-1.amazonaws.com.cn/dgl-data/dataset/openhgnn/acm4GTN.zip} shares a similar data characterises with DBLP. It contains three types of nodes (Paper(P), Author(A) and Subject(S)) and four types of edges (P-A, A-P, P-S and S-P). The papers are labelled according to three different categories. It also uses bag-of-words to construct the initial node features.
    \item \textbf{IMDB}\footnote{ https://s3.cn-north-1.amazonaws.com.cn/dgl-data/dataset/openhgnn/imdb4GTN.zip} contains movie(M), actors(A) and directors(D) as node types. The genres of movies are used as different labels for each node. Node features are also initialised using bag-of-words.
\end{itemize}

\subsection{Baselines}
To verify the effectiveness of the proposed model, we compare it with several state-of-the-art models designed for homogeneous and heterogeneous graph learning, respectively.
\begin{itemize}
    \item \textbf{GCN}~\cite{gcn} is a graph convolutional network designed specifically for homogeneous graph learning.
    \item \textbf{GAT}~\cite{gat} is the first work that introduces attention mechanism in homogeneous graph learning. It enables the weighted message aggregations from neighbor nodes. 
    \item \textbf{GraphSAGE}~\cite{graphsage} designs a sampling approach when aggregating messages from neighbor nodes. It also supports different aggregation functions. 
    \item \textbf{GraphSAINT}~\cite{graphsaint} splits nodes and edges from a bigger graph into a number of subgraphs, on which the GCN is applied for node representation learning.
    \item \textbf{HAN}~\cite{han} uses attention techniques on heterogeneous graph learning, in which the node embeddings are updated through manually designed meta-paths. 
    \item \textbf{HGT}~\cite{hgt} designs type-specific attention layers, assigning different trainable parameters to each node and edge type.
    \item \textbf{R-HGNN}~\cite{rhgnn} proposes a relation-aware method to learn the semantic representation of edges while discerning the node representations with respect to different relation types.
\end{itemize}

\subsection{Experiment Setup}
We describe the experiment setup for both baseline models and the proposed PC-HGN in this part. For baseline models dependent on attention mechanisms (i.e., GAT, HAN, and HGT), we set the number of attention heads to 8 uniformly. The sample window of GraphSAGE is set to 10~\cite{han2022openhgnn}. We
adopt the node sampling strategy for GraphSAINT and use 8000 as the node budget and 25 as the number of subgraphs (default from the released code). We use 256 as the hidden dimension of node representation throughout all baseline models for fair comparisons, and the layer depth of all models is set to 3. For the proposed PC-HGN, we set the number of kernels to 64 and pooling size to 2. We randomly split all three datasets into train/val/test with the ratio of 0.2/0.1/0.7, respectively. All models are trained with a fixed 200 epochs, using an early stopping strategy when the performance on the validation set is not improved for 50 consecutive epochs. All trainable parameters of the neural network are initialized through Xavier~\cite{glorot2010understanding} and optimized using Adam~\cite{kingma2014adam} with the learning rate 15e-4. We use both f1-micro and f1-macro as the evaluation metrics.The code of baseline models are available at https://github.com/BUPT-GAMMA/OpenHGNN.



\subsection{Performance Comparison and Analysis}
\subsubsection{Effectiveness of representation learning.} 
\begin{figure}[t!]
        \centering
        \includegraphics[width=0.4\textwidth]{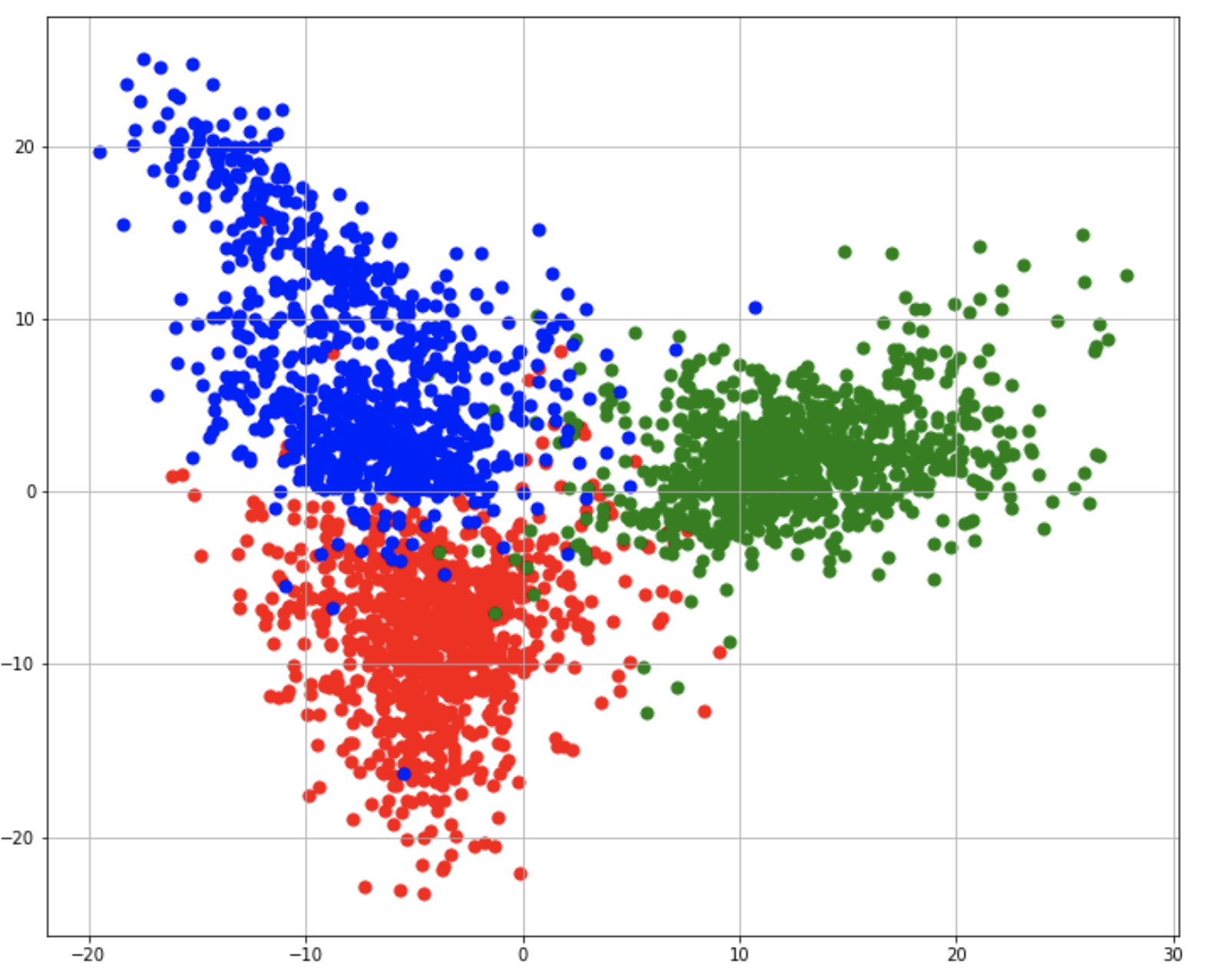}
		\caption{Paper embedding visualization of different classes}
		\label{fig:clustering}
		\vspace{-3mm}
\end{figure}

To answer Q1, we evaluate the performance of our proposed model comparing with other baselines and record the experimental results in Table~\ref{tab:performance}. It can be clearly observed that our model achieves the best result on all datasets against both homogeneous and heterogeneous representation learning methods. For traditional homogeneous graph learning models, GraphSAGE generally achieves the best result even when learning the node representations on heterogeneous graphs. Interestingly, other homogeneous graph learning models also obtain promising results compared with heterogeneous graph learning models such as HAN and HGT. Regarding the state-of-the-art heterogeneous graph learning models, R-HGNN reports the second-highest score for both F1-micro and F1-macro, implying the significance of integrating semantic information on different edge types. Furthermore, our proposed model shows a better performance than attention-based R-HGNN, which attributes to incorporating the structural heterogeneity of the graph by combing both relation-specific pooling and cross-relation convolution from neighbor nodes simultaneously. Apart from that, our model obtains a significant improvement on both metrics compared with the meta-paths dependent model HAN. We also visualize the embedding result of PC-HGN for a more intuitive view. Specifically, we use Principal Component Analysis (PCA)~\cite{PCA} to project the learnt node embeddings of paper of ACM into a 2-dimensional space. As shown in Fig.~\ref{fig:clustering}, we can see a clear division among different paper classes, implying the paper nodes has been represented properly.

\begin{figure}[t!]
        \centering
        \includegraphics[width=0.5\textwidth]{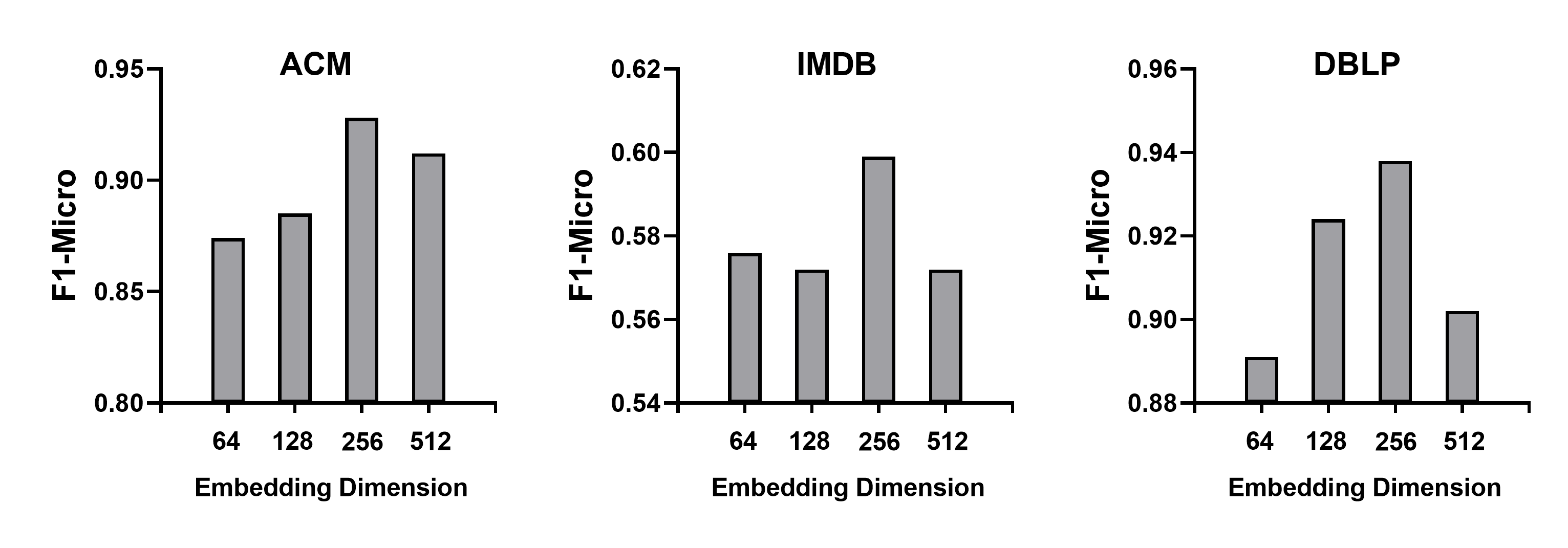}
		\caption{Impact of embedding dimensions}
		\label{fig:emd}
		\vspace{-3mm}
\end{figure}

\subsubsection{Analysis of cross-relation convolutions.} We investigate the contribution of cross-relation convolutions in this part (RQ2). Specifically, we conduct experiments to remove the filter-based convolutions and replace them with two steps: 1) mean operation on the sampled neighbors embeddings $\Tilde{\emph{X}}_{\mathcal{N}_{v}}^{t}$ to output the hidden embedding under relation $t$; 2) stacking hidden embeddings from different relations as the input to the MLP. Table~\ref{tab:ablation} provides the comparison results of PC-HGN and its varient. It can be seen a significant improvement on performance when adopting cross-relation convolutions, leading to a margin by 11.4\% on DBLP, 11.6\% on ACM and 7.1\% on IMDB, respectively.

\begin{table}[t!]
\resizebox{\linewidth}{!}{
\begin{tabular}{llll}
\cline{1-4}
\multicolumn{1}{l||}{Models}          & \multicolumn{1}{l||}{DBLP}  & \multicolumn{1}{l||}{ACM}   & IMDB   \\ \hline\hline
\multicolumn{1}{l||}{PC-HGN w/o Conv} & \multicolumn{1}{l||}{0.824} & \multicolumn{1}{l||}{0.812} & 0.528  \\ \cline{1-4}
\multicolumn{1}{l||}{PC-HGN}          & \multicolumn{1}{l||}{0.938}  & \multicolumn{1}{l||}{0.928} & 0.599  \\ \hline\hline
\end{tabular}
}
\caption{Comparison of PC-HGN with and without cross- relation convolutions on different datasets}
\label{tab:ablation}
\vspace{-4mm}
\end{table}

\subsubsection{Impact of node embedding size.} To answer RQ3, we test the impact of different node embedding sizes on the performance. As shown in Fig.~\ref{fig:emd}, it can be observed clearly that the performance of the model increases steadily at first and then starts to drop, reporting the best result with the embedding size 256. Intuitively, larger dimension sizes bring in extra redundancies in the kernel space, thus causing unexpected performance drops after convolutions.

%% file: 5_conclusion.tex
\section{Conclusion}
In this paper, we propose a novel relation-centric heterogeneous graph learning method called PC-HGN, aiming to better incorporate the information of multi-relation types without concerning pre-defined meta-paths in heterogeneous graphs. The proposed PC-HGN designs an
adaptive importance-based pooling to enable sampling for each relation type, followed by the cross-relation convolutions to incorporate the structural heterogeneity of graphs into the embedding space. The experiment results show the effectiveness of PC-HGN comparing with other baselines. Essentially, this research sheds light on the way to heterogeneous graph learning in addition to meta-path or attention-driven models.